\documentclass[10pt,twocolumn,letterpaper]{article}
\usepackage[pagenumbers]{cvpr} 
\definecolor{cvprblue}{rgb}{0.21,0.49,0.74}
\usepackage[pagebackref,breaklinks,colorlinks,allcolors=cvprblue]{hyperref}
\usepackage{float}
\usepackage{multirow}
\usepackage[marginal]{footmisc}

\def\paperID{XXXX}
\def\confName{CVPR}
\def\confYear{2025}

\newcommand{\methodname}{Layout2Scene}
\title{
    \methodname: 3D Semantic Layout Guided Scene Generation via \\Geometry and Appearance Diffusion Priors
}

\author{
Minglin Chen$^{1}$
~
Longguang Wang$^{1}$
~
Sheng Ao$^{1}$
~
Ye Zhang$^{1}$
~
Kai Xu$^{2}$
~
Yulan Guo$^{1}$\footnotemark[1]
\vspace{0.1cm}\\
{\normalsize $^{1}$ The Shenzhen Campus of Sun Yat-sen University, Sun Yat-sen University \quad $^{2}$ National University of Defense Technology}
}

\begin{document}
\twocolumn[{
\maketitle
\begin{figure}[H]
    \hsize=\textwidth
    \centering
    \begin{subfigure}{0.5\linewidth}
        \begin{minipage}[t]{\linewidth}
            \includegraphics[width=\linewidth]{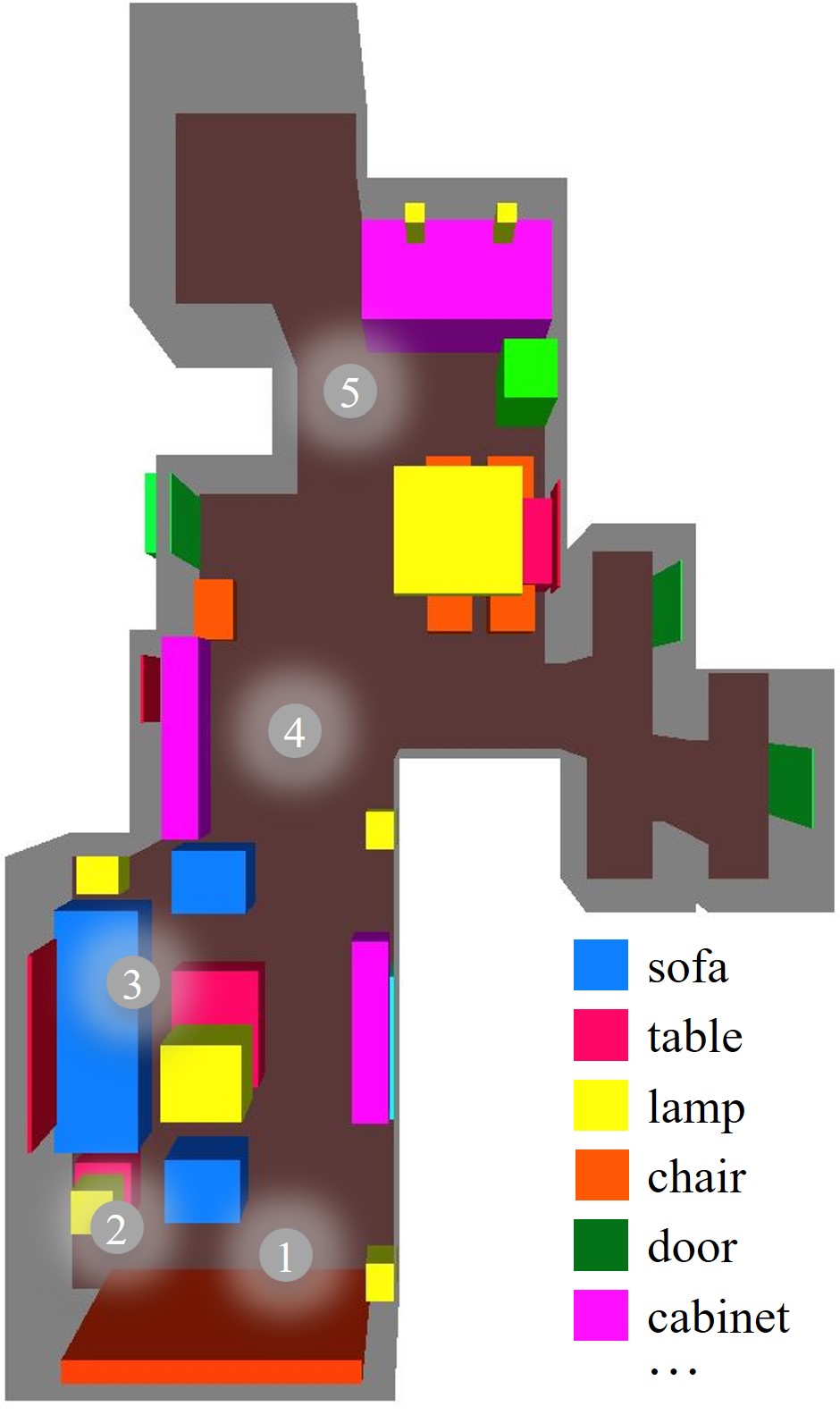}
        \end{minipage}
        \caption{3D semantic layout (input)}
    \end{subfigure}
    \begin{subfigure}{1.5\linewidth}
        \begin{minipage}[t]{\linewidth}
            \includegraphics[width=\linewidth]{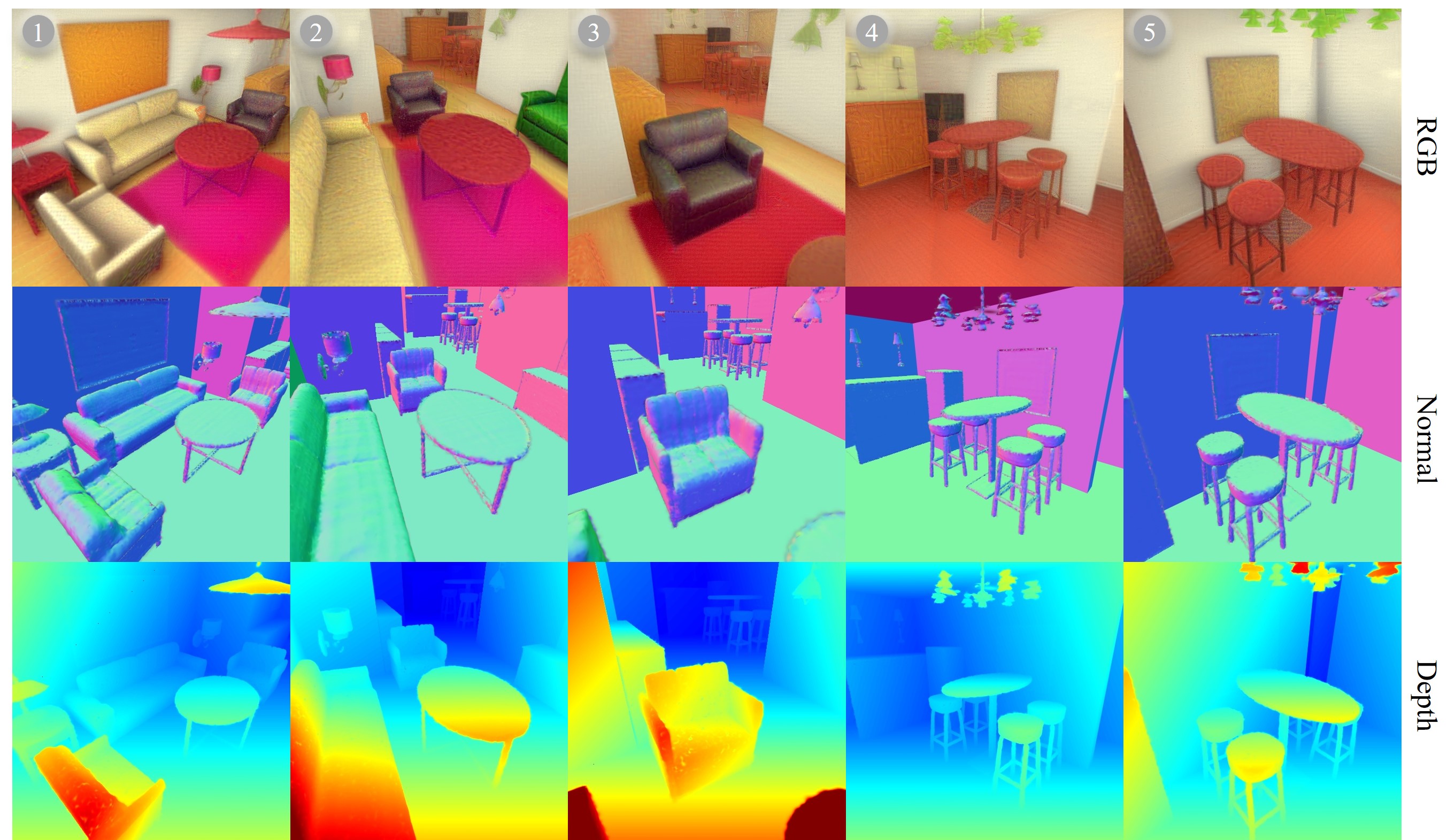}
        \end{minipage}
        \caption{Renderings of generated 3D scene}
    \end{subfigure}
    \caption{
        \textbf{\methodname} is a 3D semantic layout guided text-to-scene generative model that can create high-fidelity geometry and appearance for complex 3D scenes, while adhering to user-provided object arrangement constraints.
        (a) The inputs are a 3D semantic layout and a text prompt of the scene. The 3D semantic layout is a collection of semantic bounding boxes, while the text prompt is a brief description (\textit{“a living room"} in the case).
        (b) The generated scene exhibits a high-realness appearance and high-quality geometry, displayed through RGB, normal, and depth renderings along a navigation trajectory.
        Furthermore, the proposed method is capable of accomplishing training in \textbf{1.5 hours} and rendering at \textbf{30 FPS} on an NVIDIA V100 GPU.
    }
    \label{fig:tesser}
\end{figure}
}]

\footnotetext[1]{Corresponding author: \textcolor{magenta}{guoyulan@mail.sysu.edu.cn}}

\begin{abstract}
3D scene generation conditioned on text prompts has significantly progressed due to the development of 2D diffusion generation models.
However, the textual description of 3D scenes is inherently inaccurate and lacks fine-grained control during training, leading to implausible scene generation.
As an intuitive and feasible solution, the 3D layout allows for precise specification of object locations within the scene.
To this end, we present a text-to-scene generation method (namely, \methodname) using additional semantic layout as the prompt to inject precise control of 3D object positions.
Specifically, we first introduce a scene hybrid representation to decouple objects and backgrounds, which is initialized via a pre-trained text-to-3D model. Then, we propose a two-stage scheme to optimize the geometry and appearance of the initialized scene separately. 
To fully leverage 2D diffusion priors in geometry and appearance generation, we introduce a semantic-guided geometry diffusion model and a semantic-geometry guided diffusion model which are finetuned on a scene dataset.
Extensive experiments demonstrate that our method can generate more plausible and realistic scenes as compared to state-of-the-art approaches. Furthermore, the generated scene allows for flexible yet precise editing, thereby facilitating multiple downstream applications. 
\end{abstract}    
\section{Introduction}
3D scene assets are virtual environments that enable us to visualize, navigate, and interact as in the real world. Their applications are across a wide range of domains, \eg, 3D understanding, virtual and augmented reality, and gaming development. Creating high-fidelity 3D scenes is a time-consuming and labor-intensive procedure, typically requiring professional designers to edit the geometry and appearance of complex objects over several days or even weeks. To alleviate the burden of scene creation, 3D reconstruction~\cite{schoenberger2016sfm,schoenberger2016mvs,yuan2024dvp,yuan2024msp} provides an alternative approach that automatically reasons the underlying geometry from multi-view images or depth sensors. Among these methods, neural radiance fields (NeRF)~\cite{mildenhall2021nerf} and 3D Gaussian splatting (3DGS)~\cite{kerbl20233d,wang2024tangram} are two representative methods to model the 3D scene through inverse rendering.
More recently, 3D generation approaches have emerged as a more efficient solution for 3D scene creation from diverse types of prompts (\eg, texts and images). 

Scene generation requires prior knowledge of the 3D world. Due to the lack of large-scale 3D scene datasets, existing approaches mainly leverage 2D diffusion priors for 3D scene generation. Pioneering works~\cite{fridman2024scenescape,hollein2023text2room,zhang20243d} iteratively generate scenes from text inputs, via a text-to-image inpainting model and a monocular depth prediction model. 
Several methods~\cite{wang2023prolificdreamer,li2024dreamscene} leverage score-distillation-sampling 
 based method to generate the scene from 2D diffusion supervision. Other methods~\cite{Tang2023mvdiffusion} attempt to generate the scene via multi-view diffusion models. A complex scene requires a detailed description, however, understanding a long description is challenging for existing diffusion models.
To better understand the object relationship in the detailed text description, several methods decompose input text into a scene graph~\cite{gao2024graphdreamer} or layout~\cite{li2024dreamscene,fang2023ctrl}.
These methods generate each object using 2D diffusion priors and then compose them into a scene.

Nevertheless, existing 3D scene generation methods still face several challenges:
1) \textbf{Controllability}. Existing text-to-scene approaches synthesize the 3D scene with stochastic object locations, lacking accurate control during generation.
2) \textbf{Ambiguity}. The textual description of a scene is holistic and global, while the supervision from the diffusion model is applied to each local observed image. As a result, implausible generated scenes are produced, \eg, too many beds in a room scene generated by Text2Room~\cite{hollein2023text2room}. 
3) \textbf{Non-Editability}. Previous 3D scene generation methods employ single scene representation (\eg, mesh and implicit field) that cannot distinguish between the objects and the background. Since the objects and the background are merged in mesh representation and the inherent indivisibility of the implicit field, the generated scene cannot be edited for downstream applications.

In this paper, we propose a novel approach to generate 3D scenes from textual descriptions and semantic layouts. The textual description describes the scene type, while the semantic layout provides precise object locations and types. The semantic layout provides fine-grained control and eliminates the ambiguity from global text description during generation. We represent the 3D scene as a hybrid representation (\ie, Gaussians~\cite{kerbl20233d,huang20242d} for objects, polygons for background) by leveraging their inherent advantages for different parts.
First, we employ an efficient text-to-3D model to initialize the object in the scene. Since the initialized scene exhibits coarse geometry and inconsistent appearance, we propose to refine the scene via 2D diffusion models. Specifically, a semantic-guided geometry diffusion model is used to refine the scene geometry, while a semantic-geometry guided diffusion model is used to generate the scene appearance. To improve the generation ability for the scene domain, we trained these two diffusion models using our produced scene dataset.

The main contributions are summarized as follows:
\begin{itemize}
\item We propose a layout-guided 3D scene generation with a two-stage optimization scheme: scene geometry refinement and scene appearance generation from 2D diffusion priors.
\item We introduce semantic-guided geometry diffusion and semantic-geometry guided diffusion, which offer high-fidelity normal, depth, and image generation ability with additional conditions. 
\item Experiments show that our method can generate more plausible and realistic 3D scenes as compared to previous approaches.
\end{itemize}
\section{Related Work}
In this section, we first review text-to-3D models.
Then, we describe the prompt-based scene generation.
Finally, we highlight existing layout-guided scene generation works similar to our task.

\subsection{Text-to-3D Generation}
3D asset creation from text prompts achieves significant progress due to the development of 2D diffusion models~\cite{rombach2022high} and 3D representation~\cite{mildenhall2021nerf,kerbl20233d}. 
Pioneer work~\cite{poole2022dreamfusion} proposed a score-distillation-sampling (SDS) method to optimize the NeRF from the text-to-image diffusion model.
The 3D objects generated by SDS exhibit over-saturation, low diversity, and unrealism. 
Several follow-up works~\cite{wang2023prolificdreamer,zhuo2024vividdreamer,liang2024luciddreamer} attempt to improve SDS by inducing consistent gradient guidance. Recently, several works~\cite{yi2024gaussiandreamer,tang2023dreamgaussian} have adapted 3D Gaussians as object representation to achieve highly efficient generation.
Besides, several works~\cite{chen2023fantasia3d,lin2023magic3d,qiu2024richdreamer} investigate disentangling the geometry and appearance in the 3D generation, which separately generates the geometry and the appearance of an object. The two-stage optimization eases the generation procedure leading to high-fidelity results.
These works mainly focus on object generation and do not explore the use of two-stage optimization in scene generation.

\subsection{Prompt-based Scene Generation}
Prompt-based scene generation has been a topic of interest in recent research studies.
Hollein \textit{et al}.~\cite{hollein2023text2room} employed a 2D inpainting diffusion model and a depth estimation model to iteratively generate the scene mesh. 
Zhang \textit{et al}.~\cite{zhang2024text2nerf} used the generated image of a 2D diffusion model to optimize the NeRF-based scene.
Besides, Wang \textit{et al}.~\cite{wang2023prolificdreamer} optimized the 3D scene using 2D diffusion prior via the variational score distillation method. Tang \textit{et al}.~\cite{Tang2023mvdiffusion} proposed a multiview diffusion model to generate panoramic images from text prompts.
To better understand the fine-grained description, several works study to decompose the text before scene generation. Fang \textit{et al}.~\cite{fang2023ctrl} parsed the text into a bounding box using a pre-trained diffusion model. Recently, Li \textit{et al}.~\cite{li2024dreamscene} employed GPT-4 to decompose the scene prompt into several object prompts, which were used to generate 3D objects.
Nevertheless, prompt-based scene generation lacks fine-grained control for the scene generation, since the text prompt is inherently inaccurate to describe the object location in the scene.

\subsection{Layout-guided Scene Generation}
The layout provides an intuitive and feasible solution to control the scene generation.
Layout-guided scene generation has been gaining increasing attention recently.
Po \textit{et al}.~\cite{po2024compositional} proposed to generate each local object, and then compose them into the scene. Similarly, Cohen \textit{et al}.~\cite{cohen2023set} iteratively optimized the local object and global scene using a 2D diffusion model. These methods can only generate small scenes with a limited number of objects. Schult \textit{et al}.~\cite{schult2024controlroom3d} leveraged a semantic-conditioned diffusion model and a depth estimation model to iteratively generate the scene mesh. More recently, Yang \textit{et al}.~\cite{yang2024scenecraft} investigated generating realistic complex 3D scenes using an iterative dataset update method~\cite{haque2023instruct}. Our work is aligned with these methods and enables the generation of higher-quality scenes.
\section{Methodology}
\begin{figure*}[hbpt]
    \centering
    \includegraphics[width=\linewidth]{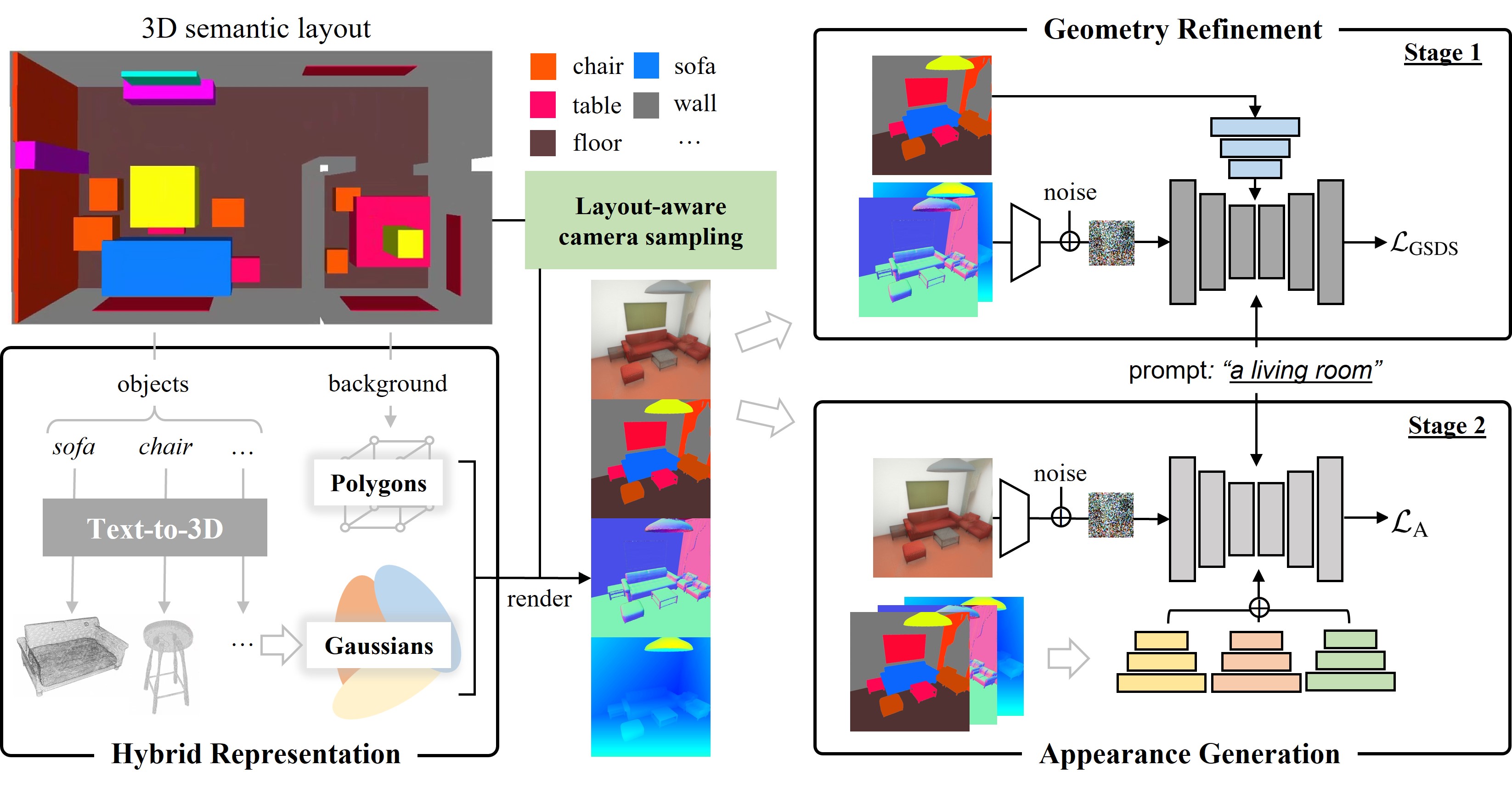}
    \vspace{-0.6cm}
    \caption{
       \textbf{Overview of \methodname}. The proposed method takes a 3D semantic layout and a prompt as input. First, we model the scene using a hybrid representation, which is initialized via a pre-trained Text-to-3D model. The layout-aware camera sampling ensures that the sampling images cover the whole scene. 
       Then, we employ a two-stage scheme to refine the geometry and appearance of the initialized scene via diffusion priors. In stage 1, we employ a semantic-guided geometry diffusion model to refine the normal and depth of the scene. In stage 2, we generate the appearance of the scene via semantic-geometry guided diffusion model. 
    }
    \label{fig:pipeline}
    \vspace{-0.4cm}
\end{figure*}
Given a 3D semantic layout along with text prompt $y$, the goal of this work is to generate a high-fidelity 3D scene model that supports novel view synthesis and scene editing. The 3D semantic layout is defined as a collection of 3D bounding boxes with textual description $\{\mathcal{B},\mathcal{T}\}$.
The 3D bounding box $\mathcal{B}$ is represented as a triplet of rotation $\mathbf{R}\in\mathbb{R}^{3\times 3}$, translation $\mathbf{t}\in\mathbb{R}^{3}$, and size $\mathbf{s}\in\mathbb{R}^{3}$, while the textual description $\mathcal{T}$ can be a single category name or a long fine-grained description.

In this section, we first introduce the scene hybrid representation that is initialized by a pre-trained text-to-3D model (Sec.~\ref{sec:SceneHybridRepresentation}).
Then, we leverage a semantic-guided geometry diffusion model to refine the scene geometry (Sec.~\ref{sec:SceneGeometryRefinement}).
Finally, we employ the semantic-geometry guided diffusion models to generate the scene appearance with an improved score distillation method (Sec.~\ref{sub:SceneAppearanceGeneration}).
Figure~\ref{fig:pipeline} illustrates the proposed method.

\subsection{Scene Hybrid Representation}
\label{sec:SceneHybridRepresentation}
A 3D scene is an environment comprising diverse categories of objects and backgrounds (\eg, wall, ground, and ceiling). 
Unlike previous approaches using single representation, we employ a hybrid representation for scene modeling, which enables the disentanglement of objects and backgrounds.
Specifically, we represent objects using 2D Gaussians~\cite{huang20242d} that are geometrically accurate, while employing explicit polygons with learnable textures for background.

\subsubsection{Object Gaussians}
We represent the object in each 3D bounding box as a set of 2D Gaussians in the Canonical space. Formally, we denote object Gaussians as a point cloud with additional properties as follows:

\begin{equation}
\label{eq:obj_gaussians}
    \mathcal{O} := \{(\mathbf{p}_i,\mathbf{\Sigma}_i,\mathbf{s}_i,\alpha_i,\mathbf{c}_i,\mathbf{t}_i)\}_i,
\end{equation}
where each primitive consists of the position $\mathbf{p}_i \in \mathbb{R}^3$, the rotation $\mathbf{\Sigma}_i\in \mathbb{R}^{3 \times 3}$, the scale $\mathbf{s}_i \in \mathbb{R}^2$, the opaque $\alpha_i \in \mathbb{R}$, and the color $\mathbf{c}_i \in \mathbb{R}^3$. Notably, we append the semantics $\mathbf{t}_i \in \mathbb{R}^3$ for each primitive, which is obtained by mapping textual description through the segmentation color protocol as used in~\cite{zhang2023adding}.

We collect all objects in the scene by transforming each object Gaussians through its corresponding 3D bounding box.
The Gaussian splatting renderer~\cite{huang20242d} is used to render the RGB image $\mathcal{I}_{o}$, the opacity map $\alpha$, the semantic map $\mathcal{S}_{o}$, the normal map $\mathcal{N}_{o}$, and the depth map $\mathcal{D}_{o}$ of the scene.

\subsubsection{Background Polygons}
The background usually has simple geometry but complex texture, \eg, a wall of the room can be approximated by a rectangle with rich patterns.
Rather than thousands of Gaussian points, we employ more efficient polygons to model background geometry as follows:

\begin{equation}
    \mathcal{BG} := \{(\mathbf{v}_j,\mathbf{e}_j)\}_j,
\end{equation}
where $\mathbf{v}_j \in \mathbb{R}^{N\times 3}$ and $\mathbf{e}_j \in \mathbb{Z}^{N\times 2}$ denote vertices and edges of polygon, respectively. Typically, $N=4$ for a wall of the room.

We employ a differential rasterizer (\ie, nvidiffrast~\cite{laine2020modular}) to render the semantic map $\mathcal{S}_b$, the normal map $\mathcal{N}_b$, and the depth map $\mathcal{D}_b$. 
The RGB image $\mathcal{I}_b$ of the background is obtained by the multiresolution hashing filed~\cite{muller2022instant}. 

The renderings of the scene are obtained by fusing the renderings of objects and background, as follows:
\begin{equation}
    \mathcal{R} = \begin{cases}
        \alpha \cdot \mathcal{R}_o + (1-\alpha) \cdot \mathcal{R}_b, & \mathcal{D}_o \leq \mathcal{D}_b \\
        \mathcal{R}_b, & \mathcal{D}_o > \mathcal{D}_b
    \end{cases}
\end{equation}
where $\mathcal{R}$ denotes the rendered RGB image $\mathcal{I}$, the semantic map $\mathcal{S}$, the normal map $\mathcal{N}$, or the depth map $\mathcal{D}$ of the scene.

\subsubsection{3D Prior Initialization}
The generation quality of Gaussians is sensitive to their initialization. Following previous works~\cite{yi2024gaussiandreamer,chen2024text} on object generation, we initialize the Gaussians using a pre-trained 3D diffusion model, \ie, Shap-E~\cite{jun2023shap}. For the object of each provided 3D bounding box, 
the pre-trained 3D diffusion model is employed to generate 3D point cloud $\{(\hat{\mathbf{p}}_i)\}_i$ via textual description $\mathcal{T}$. 
Then, the positions of object Gaussians are directly initialized using $\hat{\mathbf{p}}_i$. 

\subsubsection{Layout-aware Camera Sampling}
Existing camera sampling strategy in scene generation is inherited from object generation approaches, which randomly sample a camera on a sphere. 
However, due to the occlusion by the wall, the spherical camera sampling cannot fully cover the whole scene for complex scene structure. 

To sample meaningful cameras in the scene, the following three conditions should be satisfied: 
(1) The sampled cameras should cover the whole scene. Otherwise, only scene representations corresponding to seen parts are optimized and unsatisfactory results are produced for unseen parts. 
(2) The sampled cameras should appear outside the object, and not be close to the surface of the object or background.
(3) The sampled cameras should primarily focus on the region where the objects are more concentrated.
To meet these requirements, we introduce a simple yet efficient layout-aware camera sampling method.
Specifically, we first sample the camera positions based on the truncated signed distance field (TSDF) probability. The sampled probability $p(\cdot)$ of camera locations $x$ can be formulated as follows:
\begin{equation}
    p(x) = \text{Norm}(\text{TSDF}(x)),
\end{equation}
where $\text{Norm}(\cdot)$ is normalization operation, and the TSDF is computed using the layout.
Then, we sampled the orientation (\ie, elevation $\theta$ and azimuth $\phi$) of the camera at location $x$ as follows:
\begin{align}
    \theta &\sim \mathcal{N}(\text{mean}(\theta_i), \text{var}(\theta_i)),\\
    \phi &\sim \mathcal{N}(\text{mean}(\phi_i), \text{var}(\phi_i)),
\end{align}
where $\mathcal{N}$ denotes the normal distribution, and 
$\theta_i=\text{asin}(v_i^z)$, $\phi_i=\text{atan}(v_i^y/v_i^x)$ are the elevation and azimuth of unit vectors $v = o_i-x$ pointing from camera to the $i$-th objects, respectively. 

\subsection{Scene Geometry Refinement}
\label{sec:SceneGeometryRefinement}
Since the initialized scene presents coarse geometry, we propose to refine the scene geometry using 2D diffusion priors.
Inspired by previous works~\cite{chen2023fantasia3d,qiu2024richdreamer} on object geometry generation, a straightforward solution is to utilize a 2D diffusion model to supervise the rendered normal and depth maps of the scene.
Although this solution generates a plausible scene geometry, it fails to adhere to the provided layout constraints. This is because, the existing 2D geometry diffusion model is only conditioned on text prompts, and suffers unawareness of the layout semantics. The initialized scene geometry is compromised to the text prompts during optimization.
To satisfy layout constraints, we propose a semantic-guided geometry diffusion with the score distillation sampling method to refine the scene geometry.

\subsubsection{Semantic-guided Geometry Diffusion}
The semantic-guided geometry diffusion aims to generate a 2D geometry image conditioned on a text description and a semantic map. We utilize normal and depth maps as geometry images, since they reflect the geometry in fine-grained and coarse-grained levels.

We build the semantic-guided geometry diffusion by training a ControlNet on the pre-trained normal-depth diffusion model (ND-Diffusion)~\cite{qiu2024richdreamer}. 
The ND-Diffusion consists of a variational auto-encoder (VAE) and a U-Net as latent diffusion model (LDM), which is pre-trained on normal-depth variants of the large-scale LAION-2B dataset. 
In our semantic-guided geometry diffusion $\epsilon_\text{g}(\cdot)$, we inject semantic conditions into the input of each block of the ND-Diffusion U-Net decoder $\mathcal{D}_\text{nd}(\cdot)$, as follows:
\begin{equation}
    \epsilon_g (\hat{\mathcal{N}}, \hat{\mathcal{D}}; y, \mathcal{S}) = \mathcal{D}_\text{nd}(\{f_i^c+f_i\}_i),
\end{equation}
where, $\{f_i^c\}_i = \mathcal{E}_\text{c}(\mathcal{S})$ is a trainable U-Net encoder $\mathcal{E}_c(\cdot)$ with input semantic map $\mathcal{S}$, and $\{f_i\}_i = \mathcal{E}_\text{nd}(\hat{\mathcal{N}},\hat{\mathcal{D}};y)$ is the ND-Diffusion U-Net encoder $\mathcal{E}_\text{nd}(\cdot)$. 
$\hat{\mathcal{N}}$ and $\hat{\mathcal{D}}$ denote the VAE latent of the normal map $\mathcal{N}$ and the depth map $\mathcal{N}$, respectively. $y$ is the text description of the scene.

The semantic-guided geometry diffusion is trained on the scene image dataset (\ie the SunRGBD dataset), which provides pairs of semantic, normal, and depth images along with text description.
The training objective is as follows:

\begin{equation}
    \mathcal{L}_\text{GLDM} := \mathbb{E}_{x,y,\epsilon,t}( || \epsilon - \epsilon_\text{g}(z_t;y,t,\mathcal{S})||_2^2 ),
\end{equation}
where $z_t$ is $t$-level noisy VAE latent $x$ of normal and depth images, and $\epsilon$ is the random noise.

\subsubsection{Scene Geometry Optimization}
Given the pre-trained semantic-guided geometry diffusion, we optimize the initialized scene geometry using the score distillation sampling method.
Specifically, we render the semantic, normal, and depth images at the randomly sampled poses, then update the object Gaussians parameters using the following gradients:
\begin{equation}
    \nabla _{\theta_\text{g}} \mathcal{L}_\text{GSDS} := \mathbb{E}_{t,\epsilon}[\omega(t)(\epsilon_\text{g}(z_t;y,t,\mathcal{S})-\epsilon))\frac{\partial x}{\partial\theta_\text{g}}],
\end{equation}
where $\omega(t)$ is the weighting function, and $x$ denotes the VAE latent of normal and depth images. We only update the Gaussians geometry parameters $\theta_\text{g}=\{(\mathbf{p}_i,\mathbf{\Sigma}_i,\mathbf{s}_i,\alpha_i)\}_i$ (see Eq.\ref{eq:obj_gaussians}).

\subsection{Scene Appearance Generation}
\label{sub:SceneAppearanceGeneration}
Given a refined scene geometry, we generate its appearance via supervision from pre-trained 2D diffusion models~\cite{rombach2022high}. 
To generate an appearance that is consistent with both semantics and geometry, we introduce a semantic-geometry guided diffusion model.

\subsubsection{Semantic-Geometry Guided Diffusion}
The semantic-geometry guided diffusion controls the pre-trained Stable Diffusion with multiple conditions (\ie, semantic, normal, and depth images).
The advantage of using multiple conditions lies in eliminating the ambiguity of a single condition.

We build the semantic-geometry guided diffusion using three ControlNets and Stable Diffusion. 
Specifically, we employ three individual ControlNets to obtain the features of semantic $\mathcal{S}$, normal $\mathcal{N}$, and depth $\mathcal{D}$ maps as follows:
\begin{equation}
    \{f_i^s\}_i = \mathcal{E}_s(\mathcal{S}), \{f_i^n\}_i = \mathcal{E}_n(\mathcal{N}), \{f_i^d\}_i = \mathcal{E}_d(\mathcal{D}),
\end{equation}
where $\mathcal{E}_*(\cdot)$ is a trainable U-Net encoder. 
The semantic-geometry guided diffusion $\epsilon_a(\cdot)$ combines the outputs of ControlNets and Stable Diffusion as follows:
\begin{equation}
    \epsilon_a (\hat{\mathcal{I}}; y, \mathcal{S}, \mathcal{N}, \mathcal{D}) = \mathcal{D}_\text{sd}(\{f_i^s+f_i^n+f_i^d+f_i\}_i),
\end{equation}
where $\{f_i\}_i$ is the feature of the U-Net encoder in Stable Diffusion, and $\mathcal{D}_\text{SD}(\cdot)$ is the U-Net decoder in Stable Diffusion.

The semantic-geometry guided diffusion is trained on the scene image dataset (Sec.\ref{experiment}) with the training objective as follows:
\begin{equation}
    \mathcal{L}_\text{ALDM} := \mathbb{E}_{x,y,\epsilon,t}( || \epsilon - \epsilon_\text{g}(z_t;y,t,\mathcal{S},\mathcal{N},\mathcal{D})||_2^2 ),
\end{equation}
where $z_t$ is $t$-level noisy VAE latent $x$ of rendered RGB images, and $\epsilon$ is the random noise.

\subsubsection{Scene Appearance Optimization}
To optimize the scene appearance using the pre-trained semantic-geometry guided diffusion, we employ the invariant score distillation method~\cite{zhuo2024vividdreamer} as follows:
\begin{align}
    \nabla _{\theta_\text{a}} \mathcal{L}_\text{ISD} &:= \mathbb{E}_{t,\epsilon}[\omega(t)(\lambda (t) \delta_\text{inv}+\omega \delta_\text{cls})\frac{\partial x}{\partial\theta_\text{a}}], \\
    \delta_\text{inv} &:= \epsilon_a(z_{t-c};y,t-c) - \epsilon_a(z_t;y,t), \\
    \delta_\text{cls} &:= \epsilon_a(z_{t};y,t) - \epsilon_a(z_t;\emptyset,t),
\end{align}
where $\omega(t)$ and $\lambda(t)$ are the weighting function from time schedule~\cite{poole2022dreamfusion,zhuo2024vividdreamer}, $\omega$ denotes the classifier-free guidance value. $z_{t-c}$ is estimated from $z_{t}$ for $c$ steps using DDIM~\cite{song2020denoising}. Besides, we use a reconstruction loss $\mathcal{L}_\text{recon}$ of the rendered image $I$ and the generated image $\hat{I}$ from $z_{t-c}$. 
The total loss of scene appearance optimization is as follows:
\begin{equation}
    \mathcal{L}_\text{A} = \mathcal{L}_\text{ISD} + \gamma \mathcal{L}_\text{recon},
\end{equation}
where $\mathcal{L}_\text{recon} = ||I-\hat{I}||_2^2$, and $\gamma$ is a balance weight.
Note that, we only optimize the Gaussian appearance parameters $\{\mathbf{c}_i\}_i$ and background hashing field.

\begin{figure*}[hbpt]
    \centering
    \begin{minipage}{\textwidth}
        \centering
        “\textit{a bedroom}”
    \end{minipage}
    
    \begin{subfigure}{0.18\linewidth}
        \begin{minipage}[t]{\linewidth}
            \includegraphics[width=\linewidth]{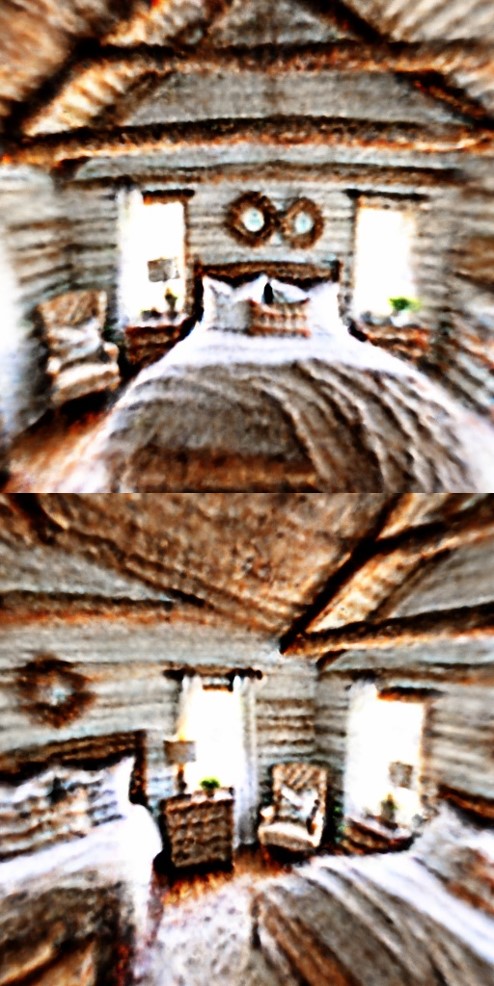}
        \end{minipage}
    \end{subfigure}
    \begin{subfigure}{0.18\linewidth}
        \begin{minipage}[t]{\linewidth}
            \includegraphics[width=\linewidth]{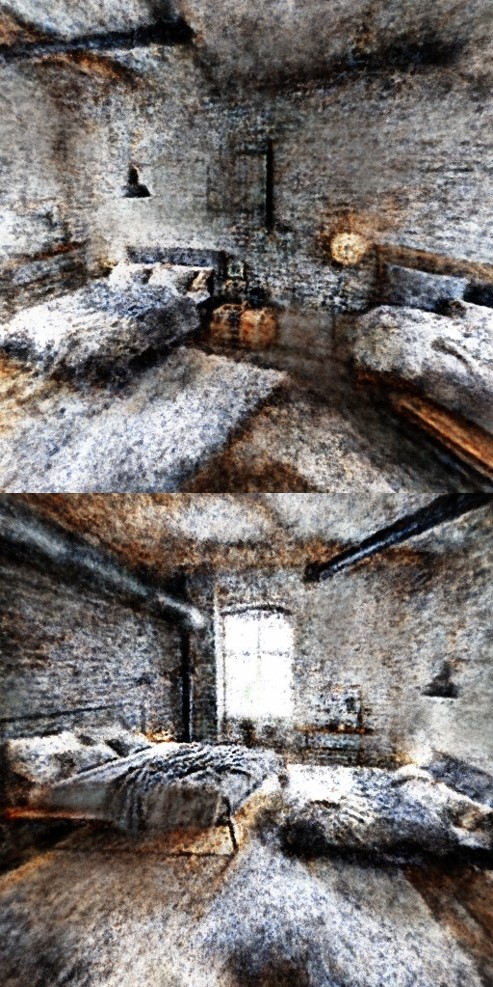}
        \end{minipage}
    \end{subfigure}
    \begin{subfigure}{0.18\linewidth}
        \begin{minipage}[t]{\linewidth}
            \includegraphics[width=\linewidth]{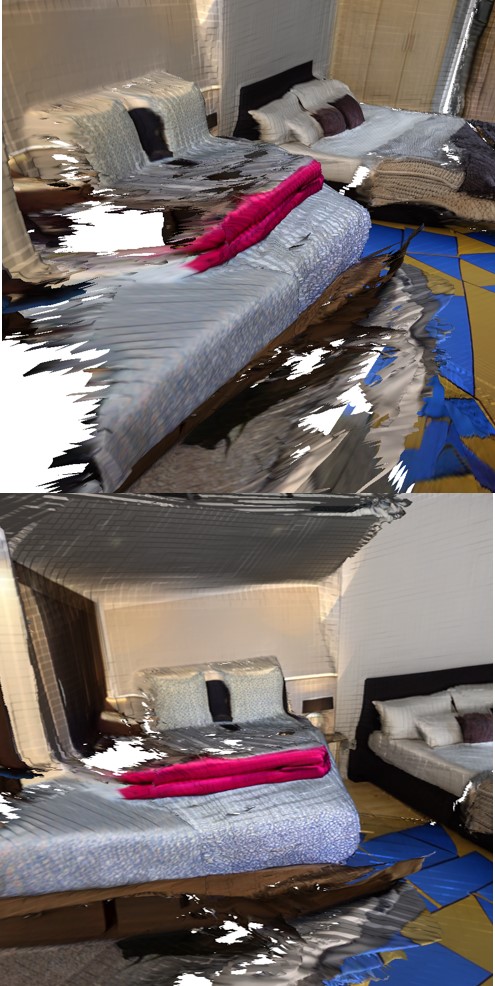}
        \end{minipage}
    \end{subfigure}
    \begin{subfigure}{0.18\linewidth}
        \begin{minipage}[t]{\linewidth}
            \includegraphics[width=\linewidth]{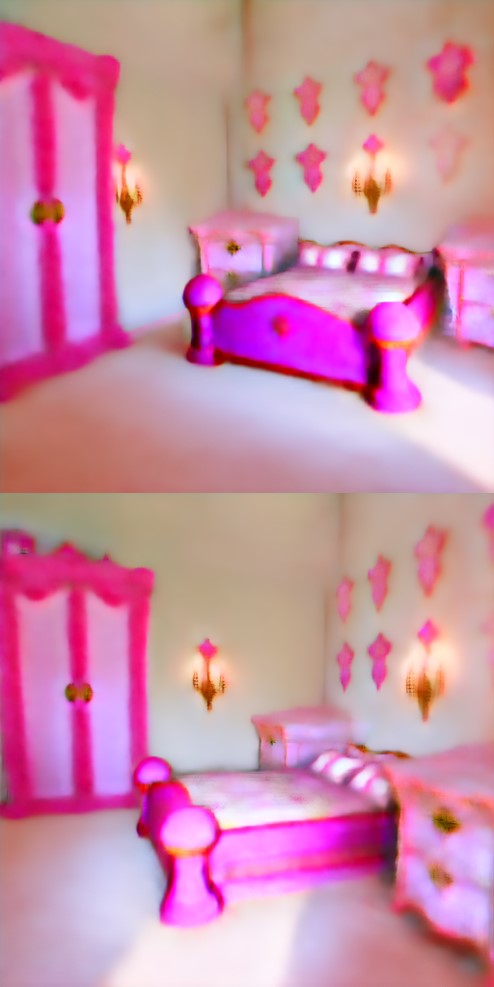}
        \end{minipage}
    \end{subfigure}
    \begin{subfigure}{0.18\linewidth}
        \begin{minipage}[t]{\linewidth}
            \includegraphics[width=\linewidth]{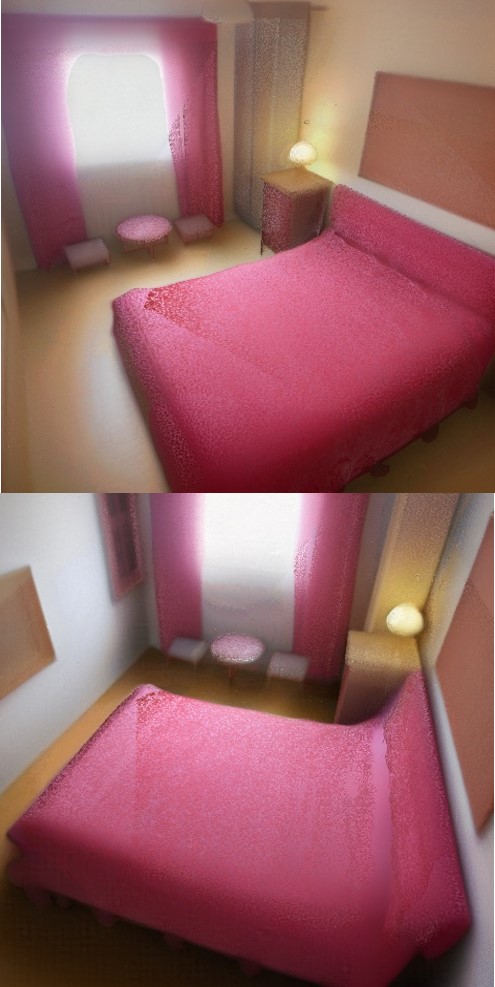}
        \end{minipage}
    \end{subfigure}

    \begin{minipage}{\textwidth}
        \centering
        “\textit{a living room}”
    \end{minipage}

    \begin{subfigure}{0.18\linewidth}
        \begin{minipage}[t]{\linewidth}
            \includegraphics[width=\linewidth]{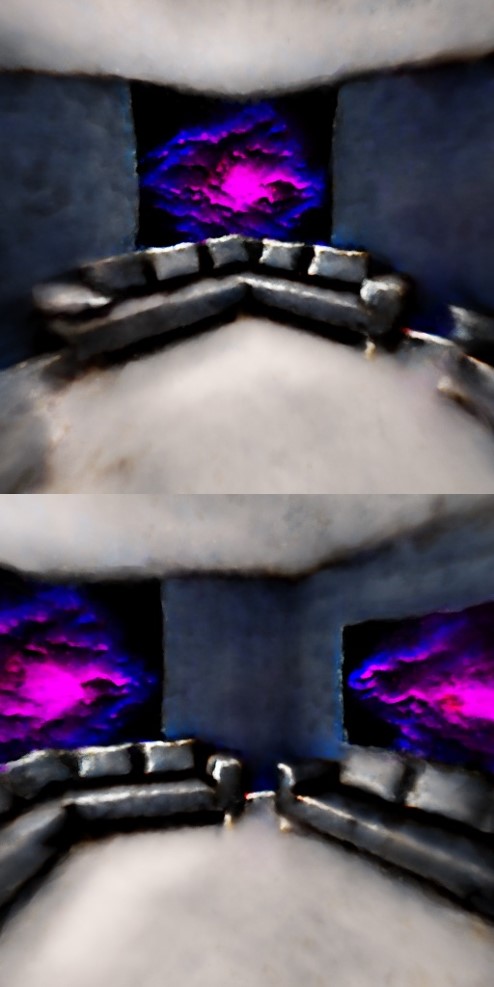}
        \end{minipage}
        \caption{DreamFusion~\cite{poole2022dreamfusion}}
    \end{subfigure}
    \begin{subfigure}{0.18\linewidth}
        \begin{minipage}[t]{\linewidth}
            \includegraphics[width=\linewidth]{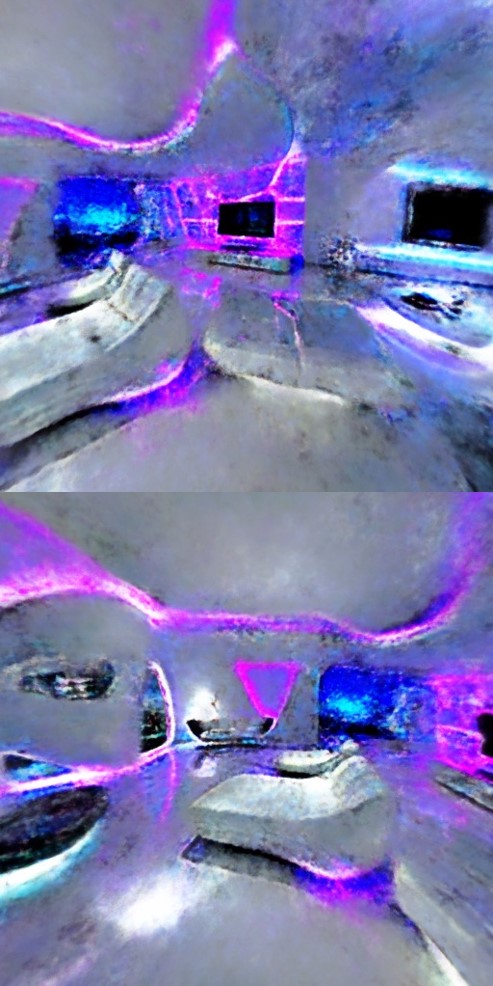}
        \end{minipage}
        \caption{ProlificDreamer~\cite{wang2023prolificdreamer}}
    \end{subfigure}
    \begin{subfigure}{0.18\linewidth}
        \begin{minipage}[t]{\linewidth}
            \includegraphics[width=\linewidth]{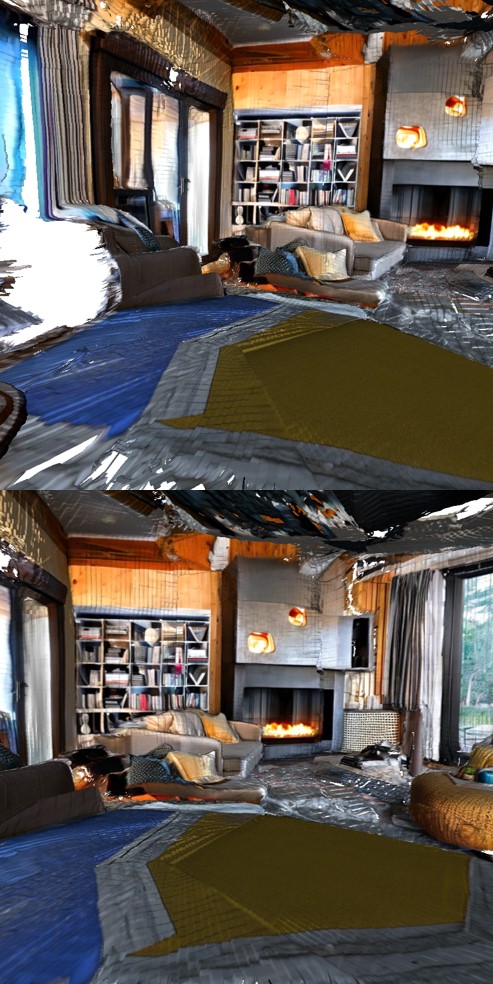}
        \end{minipage}
        \caption{Text2Room~\cite{hollein2023text2room}}
    \end{subfigure}
    \begin{subfigure}{0.18\linewidth}
        \begin{minipage}[t]{\linewidth}
            \includegraphics[width=\linewidth]{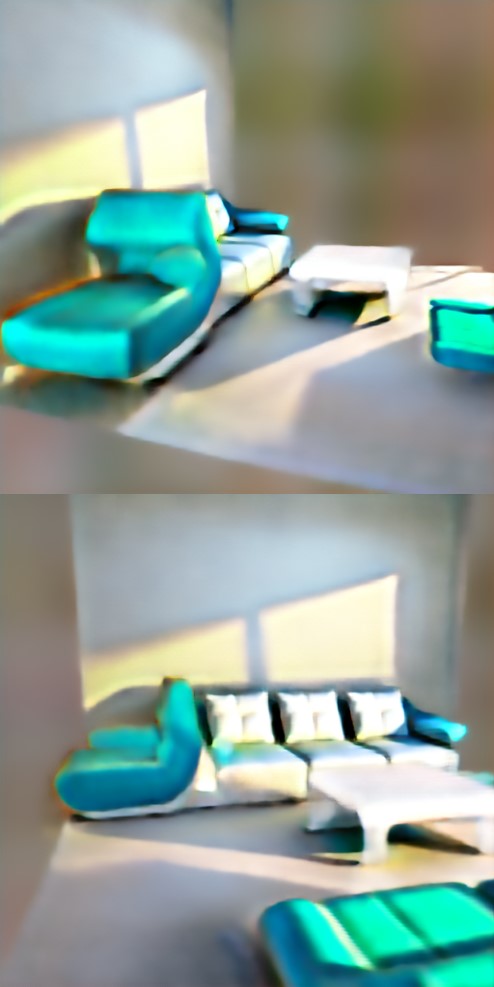}
        \end{minipage}
        \caption{Set-the-Scene~\cite{cohen2023set}}
    \end{subfigure}
    \begin{subfigure}{0.18\linewidth}
        \begin{minipage}[t]{\linewidth}
            \includegraphics[width=\linewidth]{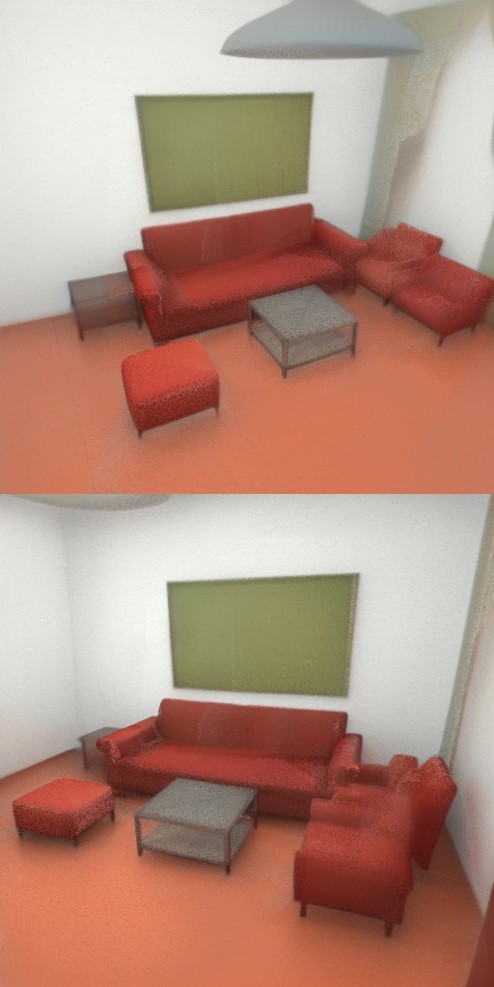}
        \end{minipage}
        \caption{Ours}
    \end{subfigure}
    \vspace{-0.3cm}
    \caption{
       {Qualitative comparisons of various scene generation approaches}.
    }
    \vspace{-0.4cm}
    \label{fig:vis_comp}
\end{figure*}

\begin{figure*}[hbpt]
    \centering
    \begin{subfigure}{\linewidth}
        \begin{minipage}[t]{\linewidth}
            \includegraphics[width=\linewidth]{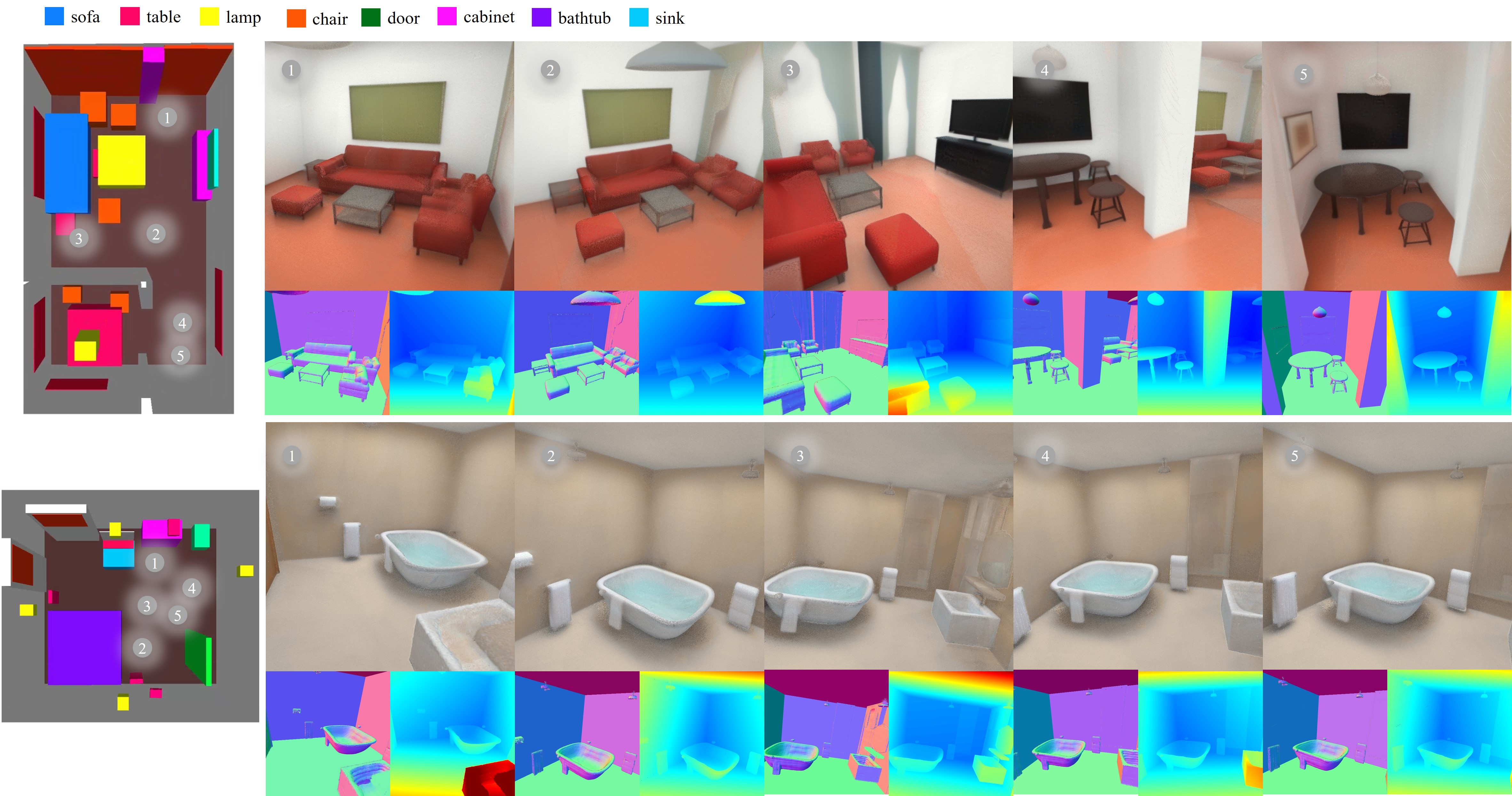}
        \end{minipage}
    \end{subfigure}
    
    \caption{
       Results of various scene types produced by the proposed method. 1st row: living room. 2nd row: bathroom. For each scene, we show the bird-eye-view of 3D semantic layout on the left, and the rendered RGB, normal, depth maps on the right.
    }
    \label{fig:layout}
    \vspace{-0.2cm}
\end{figure*}

\section{Experimental Results}
\label{experiment}

\subsection{Dataset}
We construct the scene dataset from the SunRGBD dataset~\cite{song2015sun} to train the semantic-guided geometry diffusion model and the semantic-geometry guided diffusion model. The SunRGBD dataset provides RGB images along with semantic and depth maps of more than 10,000 scenes. For the textual prompt, we employ the BLIP-2~\cite{li2023blip} model to caption the image with the question “\textit{what is the type of the scene?}". Furthermore, we use the StableNormal~\cite{ye2024stablenormal} model to estimate the normal from each RGB image.
During the training of semantic-guided geometry diffusion, we use the normal and inverse depth maps as targets, while the prompt and semantic maps as conditions. In the training of the semantic-geometry guided diffusion model, we use RGB images as the target, while the prompt, the semantic map, the semantic map, and the normal map as conditions.

\subsection{Evaluation Setup and Metrics}
We evaluate the proposed method on 20 scene layouts with various types of text prompts.
The layouts cover typical scene types (\eg, bedroom, living room, etc), and Each layout usually consists of tens of object bounding boxes.
Each bounding box is presented with a semantic category, \eg, \textit{bed}, \textit{sofa}, \textit{table} etc.
In the geometry refinement, we provide the prompt which only describes the scene type (\eg, \textit{a bedroom}).
In the appearance generation, we provide the prompt containing the scene style and type along with a fixed prefix (\eg, \textit{a DSLR photo of modern type bedroom}).
During comparisons, we set the prompt to other approaches as same as in the appearance generation.

Following previous approaches~\cite{schult2024controlroom3d}, CLIP score (CS) and Inception score (IS) are employed as metrics. For performance evaluation, we randomly rendered 120 RGB images for each scene to calculate averaged scores.

\subsection{Implementation Details}
We implemented the proposed method based on the ThreeStudio~\cite{threestudio2023} framework.
In the scene geometry refinement, we use the Adam optimizer with a learning rate of $5\times10^{-3}$, $5\times10^{-3}$, and $5\times10^{-2}$ for scaling, rotation, and opacity, respectively. The learning rate of position is initially set to $4\times10^{-3}$ and exponentially decreased to $8\times10^{-5}$ during training. 
In the appearance generation, we use the Adam optimizer with a learning rate of $5\times10^{-3}$ and $1\times10^{-2}$ for color and background, respectively. 
We optimize the scene geometry and appearance for 5000, and 10000 steps, respectively, on a single NVIDIA V100 GPU.

We trained the semantic-guided geometry diffusion model and the semantic-geometry guided diffusion model on the SunRGBD dataset for 120k steps using 4 NVIDIA V100. The batch size of each GPU is set to 16 and 2 for these two models, respectively.
The optimizer and the learning rate are the same as suggested in ControlNet.

\subsection{Comparisons with Existing Approaches}
We compare the proposed method with competitive prompt-based and layout-guided scene generation approaches. 
For prompt-based methods, we compare to DreamFusion~\cite{poole2022dreamfusion}, ProlificDreamer~\cite{wang2023prolificdreamer}, and Text2Room~\cite{hollein2023text2room}, which leverage 2D diffusion priors for scene generation.
For layout-guided methods, we compare to Set-the-Scene~\cite{cohen2023set}.

\subsubsection{Qualitative Comparisons}

Figure~\ref{fig:vis_comp} visualizes the rendered RGB images of various scene generation methods.
We found that the prompt-based approaches usually produce implausible scene structure.
For example, the bedroom consists of two beds, and the living room structure is incorrect. 
This is because the prompt-based approaches apply the diffusion prior solely to the local image, which prevents them from ensuring global consistency.
With the semantic layout constraints, our method can generate a more plausible and consistent 3D scene.

Besides, as compared to DreamFusion and ProlificDreamer, our method can render a more realistic RGB image along with high-fidelity normal and depth maps (Fig.~\ref{fig:layout}). 
The results of Text2Room contain holes in the mesh, while our hybrid representation naturally prevents this issue.
Although the Set-the-Scene methods can be guided by 3D layout, their renderings are blurred and contain floater artifacts. In contrast, our method can synthesize more clean RGB images, due to the disentanglement of geometry and appearance during optimization.

\begin{table}[htbp]
  \centering
  \footnotesize
  \caption{Quantitative comparisons of scene generation approaches. (Note that Tr.Time and FPS denote the training times and the number of frames per second in rendering, respectively.)}
    \setlength{\tabcolsep}{3.0mm}{
        \begin{tabular}{l|cc|cc}
        \toprule
              & CS$\uparrow$   & IS$\uparrow$    & Tr.Time$\downarrow$ & FPS$\uparrow$ \\
        \midrule
        \multicolumn{5}{l}{Prompt-based scene generation} \\
        \midrule
        DreamFusion~\cite{poole2022dreamfusion} & 18.92 & 2.31  & 1 hour & 1.4 \\
        ProlificDreamer~\cite{wang2023prolificdreamer} & 18.51 & 2.19  & 2 hours & 2.3 \\
        Text2Room~\cite{hollein2023text2room} & 22.42 & 3.21  & \textbf{0.3 hours} & - \\
        \midrule
        \multicolumn{5}{l}{Layout-guided scene generation} \\
        \midrule
        Set-the-Scene~\cite{cohen2023set} & 19.24 & 2.77  & 4 hours & 0.3 \\
        Ours  & \textbf{25.69} & \textbf{3.51}  & 1.5 hours & \textbf{30} \\
        \bottomrule
        \end{tabular}%
    }
  \label{tab:quantitative_comparisons}%
  \vspace{-0.4cm}
\end{table}%

Figure~\ref{fig:layout} and Figure~\ref{fig:tesser} show the results of the proposed method on various types of layouts.
As compared to Set-the-Scene, the proposed method only uses object bounding boxes instead of fine-grained voxels as input. 
This can reduce the requirements for input, making the control of scene generation simpler.
Even with more simple input, our method produces more clean scene. 

\subsubsection{Quantitative comparisons}
Table~\ref{tab:quantitative_comparisons} presents the quantitative comparisons of our method against baselines in scene generation. Among these methods, the proposed method supports layout guidance and achieves high-quality renderings.
Our method achieves a CS of 25.69 and an IS of 3.51, outperforming state-of-the-art Set-the-Scene by 6.45 and 0.74, respectively. This indicates that the generated scenes by our method are of higher fidelity to the prompt and more realistic.

Notably, our method also exhibits high efficiency in training and rendering of layout-guided scene generation. 
The training of each scene takes approximately 1.5 hours (a few minutes for initialization, 0.5 hours for geometry refinement, and 1 hour for appearance generation) on a single NVIDIA V100 GPU. 

\subsection{Ablation Study}

\subsubsection{Effect of Geometry Prior}
\begin{figure}[t]
    \centering
    \begin{subfigure}{0.48\linewidth}
        \begin{minipage}[t]{\linewidth}
            \includegraphics[width=\linewidth]{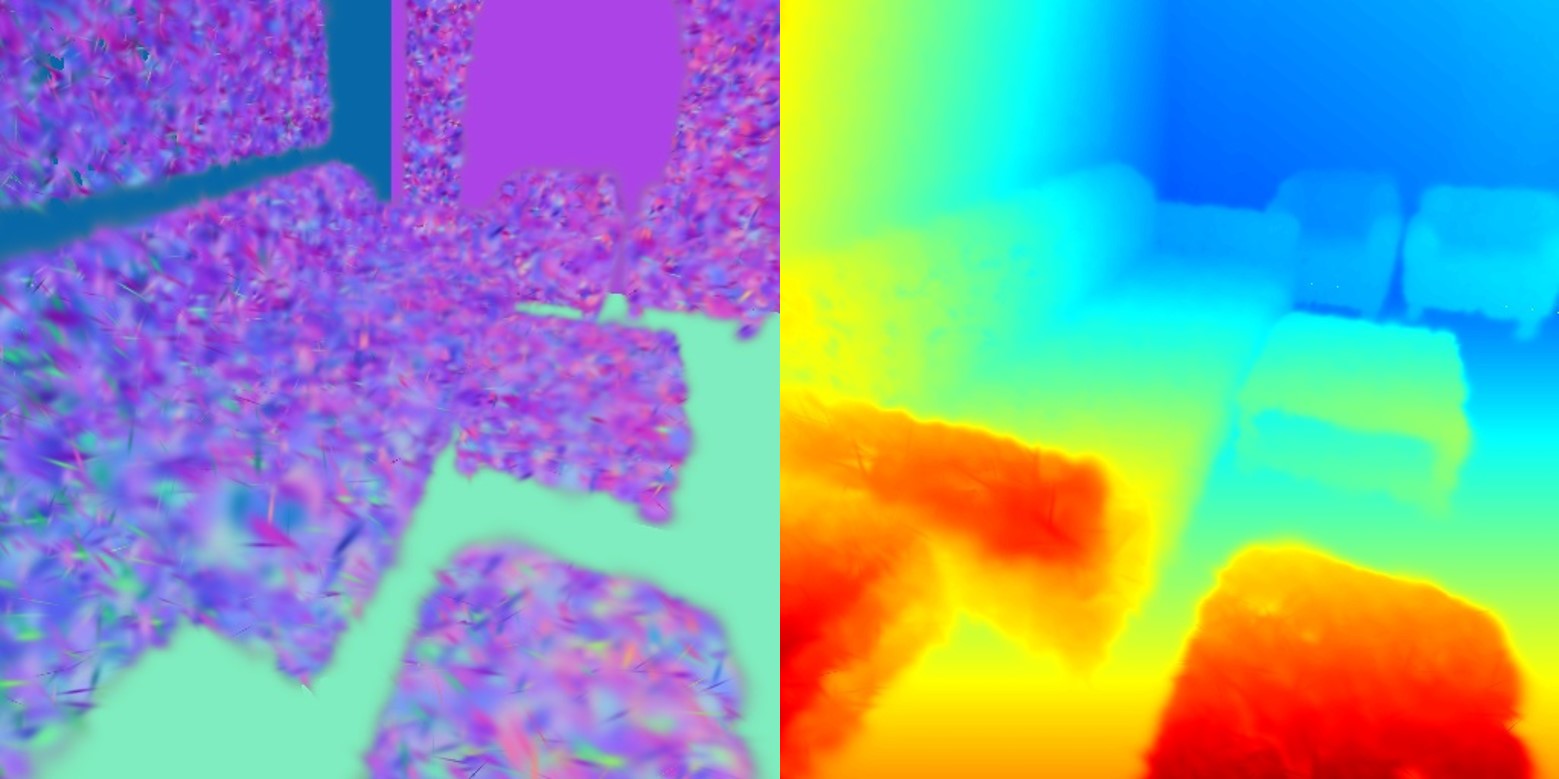}
        \end{minipage}
        \caption{w/o geometry diffusion prior}
    \end{subfigure}
    \begin{subfigure}{0.48\linewidth}
        \begin{minipage}[t]{\linewidth}
            \includegraphics[width=\linewidth]{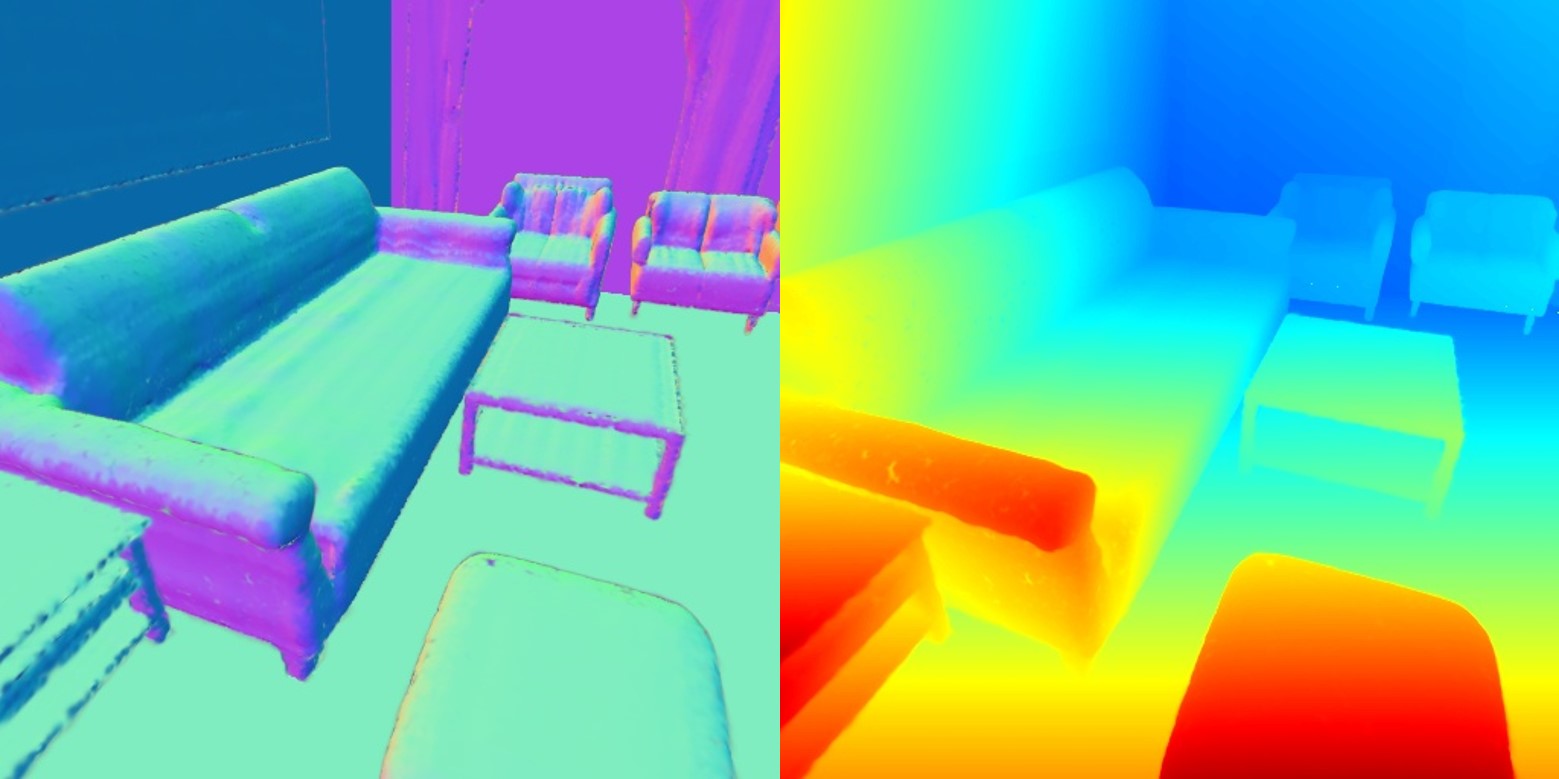}
        \end{minipage}
        \caption{w/ geometry diffusion prior}
    \end{subfigure}
    
    \caption{
        Visualization of normal and depth maps with and without the geometry diffusion prior.
    }
    \label{fig:ablation_geo}
    \vspace{-0.4cm}
\end{figure}

To verify the effectiveness of semantic-guided geometry diffusion, we visualize the normal and depth maps with and without the geometry diffusion prior.
It can be observed from Fig.~\ref{fig:ablation_geo}  that the geometry diffusion significantly refines the normal of the scene with finer details.

\subsubsection{Effect of Appearance Prior}
\begin{figure}[hbpt]
    \centering
    \begin{subfigure}{0.3\linewidth}
        \begin{minipage}[t]{\linewidth}
            \includegraphics[width=\linewidth]{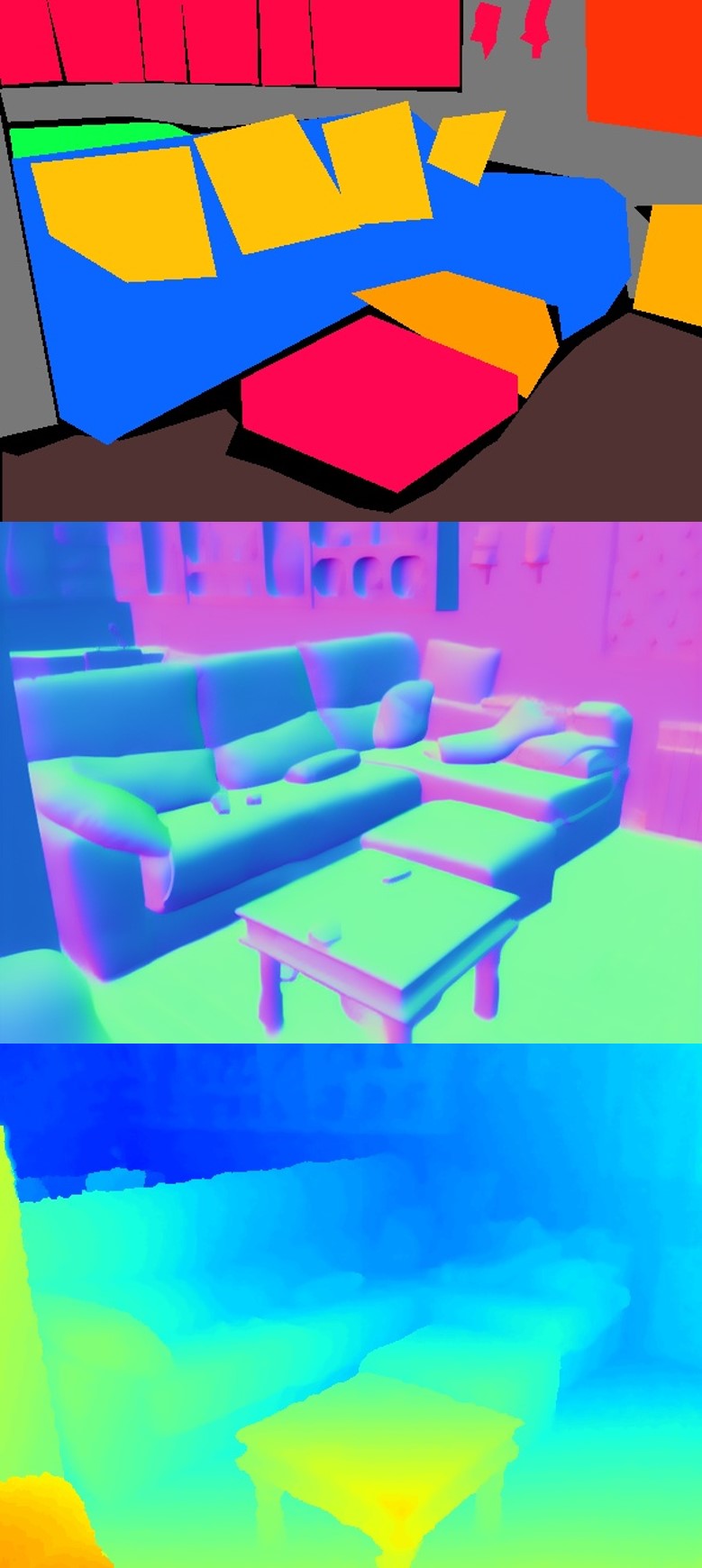}
        \end{minipage}
        \caption{Input}
    \end{subfigure}
    \begin{subfigure}{0.3\linewidth}
        \begin{minipage}[t]{\linewidth}
            \includegraphics[width=\linewidth]{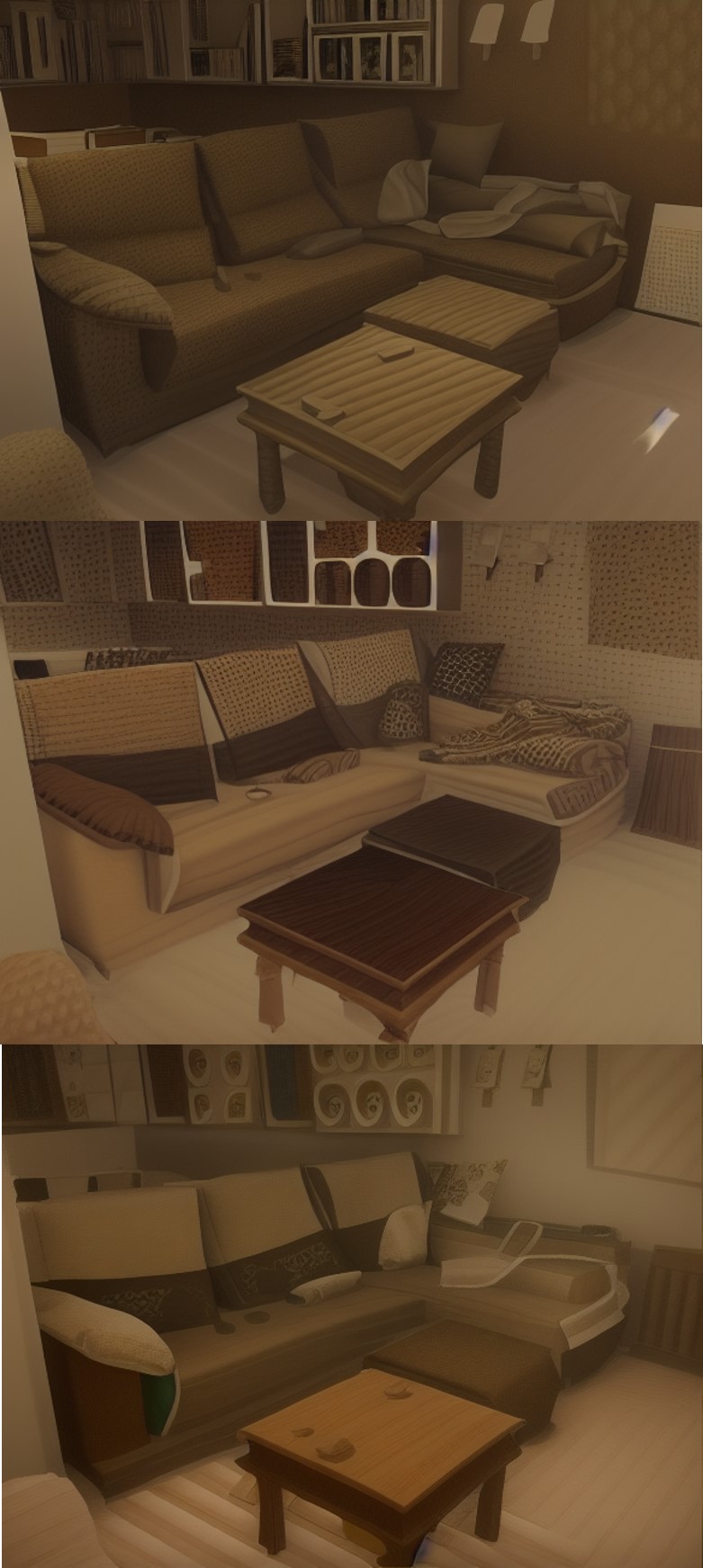}
        \end{minipage}
        \caption{ControlNet~\cite{zhang2023adding}}
    \end{subfigure}
    \begin{subfigure}{0.3\linewidth}
        \begin{minipage}[t]{\linewidth}
            \includegraphics[width=\linewidth]{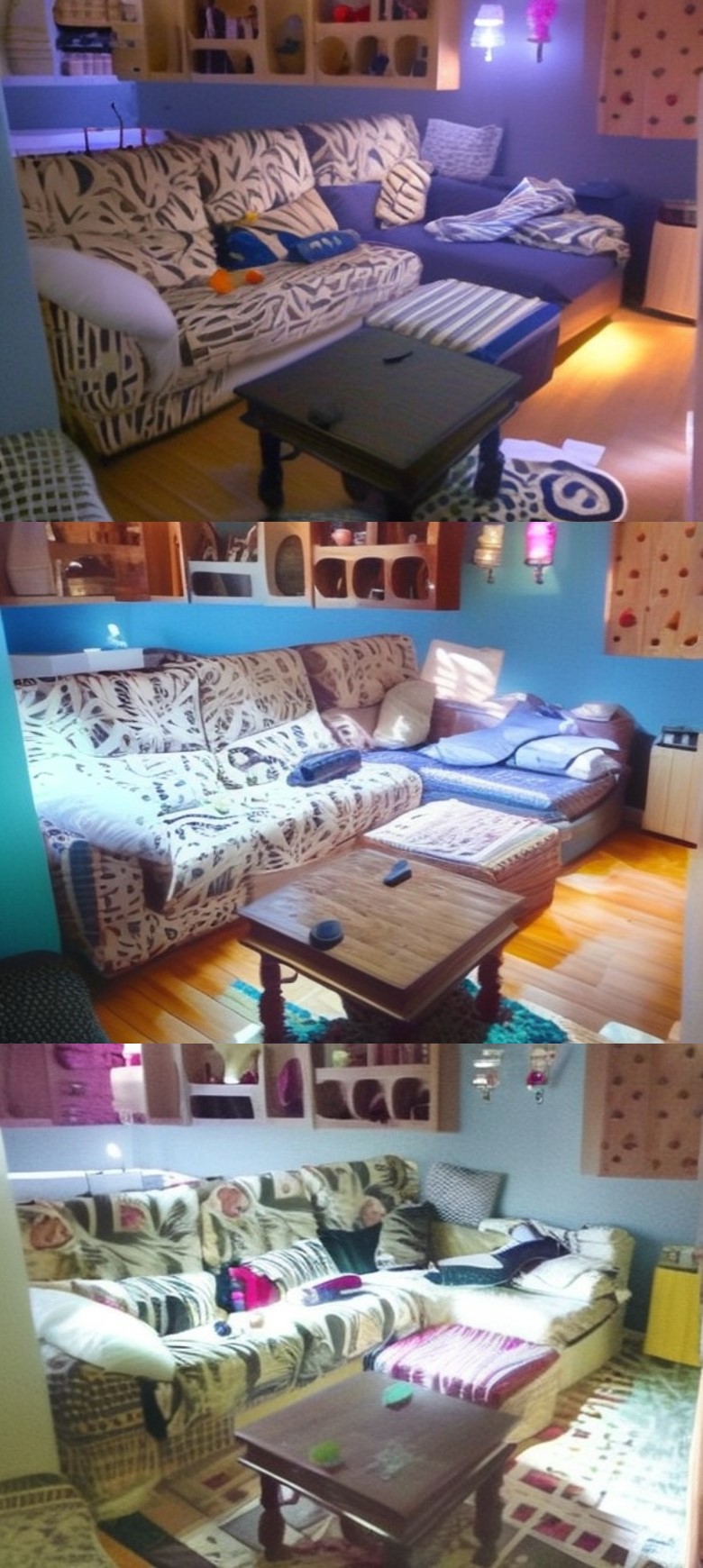}
        \end{minipage}
        \caption{Ours}
    \end{subfigure}
    
    \caption{
        Comparisons of generated scene images under multiple conditions.
    }
    \label{fig:ablation_appearance}
    \vspace{-0.4cm}
\end{figure}

We conduct experiments to study the effectiveness of appearance prior introduced by the semantic-geometry guided diffusion model. As a comparison, we employ the original ControlNet to generate the image with multiple conditions (\ie, semantic, normal, and depth maps). As shown in Fig.~\ref{fig:ablation_appearance}, the images generated by ControlNet suffer limited diversity and are unrealistic. With our semantic-geometry guided diffusion model, our method synthesizes scene images with high diversity and realness, especially for the realistic lightning effects.

\section{Conclusion}
In this paper, we present a 3D semantic layout guided text-to-scene generation method.
The 3D scene is modeled as a hybrid representation which is initialized via a pre-trained text-to-3D model. We optimize the initialized scene presentation via a two-stage scheme. Specifically, a semantic-guided geometry diffusion model is first employed for scene geometry refinement, and then a semantic-geometry guided diffusion model is adopted for scene appearance generation. To fully leverage 2D diffusion priors in geometry and appearance generation, we trained these two diffusion models on a scene dataset. Extensive experiments show that the proposed method can generate more plausible and realistic 3D scenes under the control of the user-provided 3D semantic layout.

{
    \small
    \bibliographystyle{ieeenat_fullname}
    \bibliography{main}

\end{document}